\documentclass{article} 
\usepackage{iclr2015,times}
\usepackage{hyperref}
\usepackage{url}
\usepackage{graphicx}
\usepackage{bm}
\usepackage{amsmath}
\usepackage{amsfonts}
\usepackage{array}
\usepackage{titlesec}

\titlespacing*{\section}{0pt}{1pt}{1pt}

\def\eps{{\epsilon}}

\def\sign{{\text{sign}}}

\def\vbeta{{\bm{\beta}}}

\def\veta{{\bm{\eta}}}

\def\vtheta{{\bm{\theta}}}

\def\vw{{\bm{w}}}
\def\vx{{\bm{x}}}

\def\mI{{\bm{I}}}

\title{Explaining and Harnessing\\ Adversarial Examples}
\author{
Ian J. Goodfellow, Jonathon Shlens \& Christian Szegedy \\
Google Inc., Mountain View, CA\\
\texttt{\{goodfellow,shlens,szegedy\}@google.com} \\
}

\iclrfinalcopy 
\iclrconference 
\begin{document}
\maketitle
\begin{abstract}
    \vspace{-.1in}
Several machine learning models, including neural networks, consistently
misclassify {\em adversarial examples}---inputs formed
by applying small but intentionally worst-case perturbations to examples from the dataset,
such that the perturbed input results in the model outputting an incorrect
answer with high confidence.
Early attempts at explaining this phenomenon focused on nonlinearity and overfitting.
We argue instead that the primary cause of neural networks' vulnerability to adversarial
perturbation is their linear nature.
This explanation is supported by new quantitative results while
giving the first explanation of the most intriguing fact about
them: their generalization across architectures and training sets.
Moreover, this view yields a simple and fast method of generating adversarial examples.
Using this approach to provide examples for adversarial training, we reduce the test
set error of a maxout network on the MNIST dataset.
\vspace{-.05in}
\end{abstract}
\vspace{-.1in}
\section{Introduction}
\citet{Szegedy-ICLR2014} made an intriguing discovery: several machine learning models,
including state-of-the-art neural networks, are vulnerable to {\em adversarial examples}.
That is, these machine learning models misclassify examples that are only slightly different
from correctly classified examples drawn from the data distribution.
In many cases, a wide variety of models with different architectures trained
on different subsets of the training data misclassify the same adversarial example.
This suggests that adversarial examples expose
fundamental blind spots in our training algorithms.

The cause of these adversarial examples was a mystery, and speculative explanations have
suggested it is due to extreme nonlinearity of deep neural networks, perhaps combined
with insufficient model averaging and insufficient regularization of the purely supervised
learning problem. We show that these speculative hypotheses are unnecessary.
Linear behavior in high-dimensional spaces is sufficient to cause adversarial examples.
This view enables us to design a fast method of generating adversarial examples that
makes adversarial training practical. We show that adversarial training can provide
an additional regularization benefit beyond that provided by using dropout~\citep{dropout} alone.
Generic regularization strategies such as dropout, pretraining,
and model averaging do not confer a significant reduction in a model's vulnerability to
adversarial examples, but changing to nonlinear model families such as RBF networks can do so.

Our explanation suggests a fundamental tension between designing models that are easy to train due
to their linearity and designing models that use nonlinear effects to resist adversarial
perturbation. In the long run, it may be possible to escape this tradeoff by designing more
powerful optimization methods that can succesfully train more nonlinear models.

\section{Related work}

~\citet{Szegedy-ICLR2014} demonstrated a variety of intriguing properties of neural networks
and related models. Those most relevant to this paper include:
\begin{itemize}
\item Box-constrained L-BFGS can reliably find adversarial examples.
\item On some datasets, such as ImageNet~\citep{ImageNet}, the adversarial examples
    were so close to the original examples that the differences were indistinguishable
    to the human eye.
\item The same adversarial example is often misclassified by a variety of 
      classifiers with different architectures or trained on different subsets of the training data.
\item 
      Shallow softmax regression models are also vulnerable to adversarial examples.
\item Training on adversarial examples can regularize the model---however, this was not
      practical at the time due to the need for expensive constrained optimization 
      in the inner loop.
\end{itemize}

These results suggest that classifiers based on modern machine learning techniques, even
those that obtain excellent performance on the test set, are not learning the true underlying
concepts that determine the correct output label. Instead, these algorithms
have built a Potemkin village that works well on naturally occuring data, but is exposed as
a fake when one visits points in space that do not have high probability in the data distribution.
This is particularly disappointing because a popular approach in computer vision is to use
convolutional network features as a space where Euclidean distance approximates perceptual distance.
This resemblance is clearly flawed if images that have an immeasurably small perceptual distance
correspond to completely different classes in the network's representation.

These results have often been interpreted as being a flaw in deep networks in particular, even
though linear classifiers have the same problem. We regard the knowledge of this flaw as an opportunity
to fix it. Indeed,
~\citet{Luca} and ~\citet{causal}
have already begun the first steps toward designing models that resist adversarial perturbation,
though no model has yet succesfully done so while maintaining state of the art accuracy on
clean inputs.

\section{The linear explanation of adversarial examples}

We start with explaining the existence of adversarial examples for linear models.

In many problems, the precision of an individual input feature is limited. For example, 
digital images often use only
8 bits per pixel so they discard all information below $1 / 255$ of the dynamic range.
 Because
 the precision of the features is limited, it is not rational for the classifier to
 respond differently to an input $\vx$ than to an adversarial input $\tilde{\vx} = \vx + \veta$
if every element of the perturbation $\veta$ is smaller than the precision of the features. Formally,
for problems with well-separated classes,
we expect the classifier to assign the same class to $\vx$ and $\tilde{\vx}$ so long as
$||\veta||_\infty < \eps$, where $\eps$ is small enough to be discarded by the
sensor or data storage apparatus associated with our problem.

Consider the dot product between a weight vector $\vw$ and an adversarial example $\tilde{\vx}$:
\[ \vw^\top \tilde{\vx} = \vw^\top \vx + \vw^\top \veta. \]
The adversarial perturbation causes the activation to grow by $\vw^\top \veta$.
We can maximize this increase subject to the max norm constraint on $\veta$ by assigning
$\eta = \text{sign}(\vw)$. If $\vw$ has $n$ dimensions and the average magnitude of an element
of the weight vector is $m$, then the activation will grow by $\eps m n$.
%
Since $||\eta||_\infty$ does not grow with the dimensionality of the problem but the change in activation
caused by perturbation by $\eta$ can grow linearly with $n$, then for high dimensional problems, we can make many
infinitesimal changes to the input that add up to one large change to the output.
We can think of this as a sort of ``accidental steganography,'' where a linear model is forced to attend
exclusively to the signal that aligns most closely with its weights, even if multiple signals are
present and other signals have much greater amplitude.

This explanation shows that a simple linear model can have adversarial examples if its input has
sufficient dimensionality. Previous explanations for adversarial examples invoked hypothesized
properties of neural networks, such as their supposed highly non-linear nature.
Our hypothesis based on linearity is simpler, and can also explain why softmax regression is vulnerable
to adversarial examples.


\section{Linear perturbation of non-linear models}

The linear view of adversarial examples suggests a fast way of generating them.
We hypothesize that neural networks are too linear to resist linear adversarial
perturbation.
LSTMs~\citep{lstm}, ReLUs~\citep{Jarrett-ICCV2009,Glorot+al-AI-2011},
and maxout networks~\citep{Goodfellow-et-al-ICML2013} are all intentionally
designed to behave in very linear ways, so that they are easier to optimize.
More nonlinear models such as sigmoid networks are carefully tuned to spend
most of their time in the non-saturating, more linear regime for the same reason.
This linear behavior suggests that cheap, analytical perturbations of a linear model
should also damage neural networks.

Let $\vtheta$ be the parameters of a model, $\vx$ the input to the model, $y$ the targets associated with
$\vx$ (for machine learning tasks that have targets) and $J(\vtheta, \vx, y)$ be the cost used to train the
neural network. We can linearize the cost function around the current value of $\vtheta$, obtaining
an optimal max-norm constrained pertubation of
\[ \veta = \eps \sign \left( \nabla_\vx J(\vtheta, \vx, y) \right). \]
We refer to this as the ``fast gradient sign method'' of generating adversarial examples.
Note that the required gradient can be computed efficiently using backpropagation.

\begin{figure}[t]
    \vspace{-.05in}
    \centering
\begin{tabular}{>{\centering\arraybackslash}m{.2\textwidth}m{.5in}>{\centering\arraybackslash}m{.2\textwidth}m{.1in}>{\centering\arraybackslash}m{.2\textwidth}}
    \centering\arraybackslash
    \includegraphics[width=.2\textwidth]{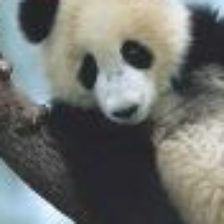} &%
    \centering\arraybackslash%
$\ +\ .007\ \times$ &%
    \includegraphics[width=.2\textwidth]{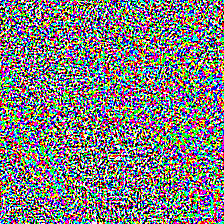} &%
    $=$ & %
    \includegraphics[width=.2\textwidth]{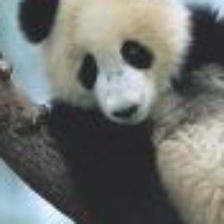} \\
    $\centering \vx$     &%
    & $\sign (\nabla_\vx J(\vtheta, \vx, y) )$ & & $\vx + \epsilon \sign (\nabla_\vx J(\vtheta, \vx, y) )$ \\
    ``panda'' &                & ``nematode''     &   & ``gibbon'' \\
    57.7\% confidence &        &   8.2\% confidence & & 99.3 \% confidence
\end{tabular}
    \caption[Fast adversarial sample generation]{
    A demonstration of fast adversarial example generation applied to GoogLeNet
    \citep{Szegedy-et-al-arxiv2014}
    on ImageNet.
    By adding an imperceptibly small vector whose elements are equal to the sign of the elements
    of the gradient of the cost function with respect to the input, we can
    change GoogLeNet's classification of the image. Here our $\eps$ of .007
    corresponds to the magnitude of the smallest bit of an 8 bit image encoding
    after GoogLeNet's conversion to real numbers.
    }
\label{panda}
\end{figure}

We find that this method reliably causes a wide variety of models to misclassify their input.
See Fig.~\ref{panda} for a demonstration on ImageNet.
We find that using $\eps=.25$, we cause a shallow softmax classifier to have an error rate of 99.9\% with an average confidence of
79.3\% on the MNIST~\citep{LeCun+98} test set\footnote{This is using MNIST pixel values in the interval [0, 1]. MNIST data
does contain values other than 0 or 1, but the images are essentially binary. Each pixel roughly encodes
``ink'' or ``no ink''. This justifies expecting the classifier to be able to handle perturbations
within a range of width 0.5, and indeed human observers can read such images without difficulty.
}. In the same setting, a maxout network misclassifies 89.4\%
of our adversarial examples with an average confidence of 97.6\%.
Similarly, using $\eps=.1$, we obtain an error rate of 87.15\% and an average probability of
96.6\% assigned to the incorrect labels
when using
a convolutional maxout network on a preprocessed version of the CIFAR-10~\citep{KrizhevskyHinton2009} test set\footnote{
    See \url{https://github.com/lisa-lab/pylearn2/tree/master/pylearn2/scripts/papers/maxout}.
    for the preprocessing code, which yields a standard deviation of roughly 0.5.
}. Other simple methods of generating adversarial examples are possible. For example, we also found that rotating
$\vx$ by a small angle in the direction of the gradient reliably produces adversarial examples.

The fact that these simple, cheap algorithms are able to generate misclassified examples serves as evidence in favor of our interpretation of adversarial
examples as a result of linearity. The algorithms are also useful as a way of speeding up adversarial training or
even just analysis of trained networks.

\section{Adversarial training of linear models versus weight decay}

\begin{figure}
    \vspace{-.05in}
    \centering
    \begin{tabular}{cccc}
        \includegraphics[width=.5in]{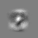}    &
        \includegraphics[width=.5in]{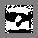} &
        \includegraphics[width=1.5in]{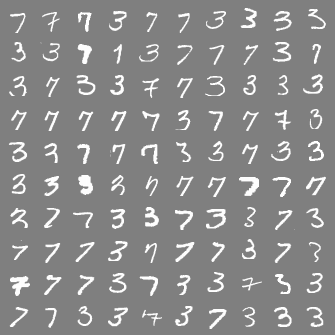}&
        \includegraphics[width=1.5in]{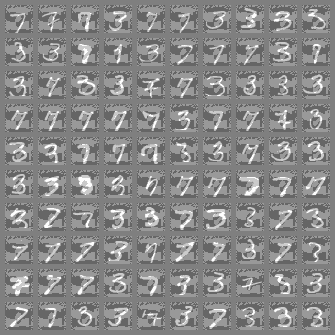}    \\
            (a) & (b) & (c) & (d)
    \end{tabular}

\caption{
    The fast gradient sign method applied to logistic regression (where it is not an approximation,
    but truly the most damaging adversarial example in the max norm box).
    a) The weights of a logistic regression model trained on MNIST.
    b) The sign of the weights of a logistic regression model trained on MNIST. This is the optimal
    perturbation. Even though the model has low capacity and is fit well, this perturbation is not
    readily recognizable to a human observer as having anything to do with the relationship between
    3s and 7s.
    c) MNIST 3s and 7s. The logistic regression model has a 1.6\% error rate on the 3 versus 7 discrimination task on these examples.
    d) Fast gradient sign adversarial examples for the logistic regression model with $\eps=.25$.
        The logistic regression model has an error rate of 99\% on these examples.
}
\label{fig:logistic_adv}
\end{figure}

Perhaps the simplest possible model we can consider is logistic regression. In this case, the fast
gradient sign method is exact. We can use this case to gain some intuition for how adversarial examples
are generated in a simple setting. See Fig.~\ref{fig:logistic_adv} for instructive images.

If we train a single model
to recognize labels $y \in \{ -1, 1\}$ with $P(y=1) = \sigma\left( \vw^\top \vx + b\right)$ where $\sigma(z)$
is the logistic sigmoid function, then training consists of gradient descent on
\[
    \mathbb{E}_{\vx, y \sim p_\text{data}} \zeta( -y (\vw^\top \vx + b))
\]
where $\zeta(z) = \log \left(1 + \exp(z) \right)$ is the softplus function.
We can derive a simple analytical form for training on the worst-case adversarial perturbation of
$\vx$ rather than $\vx$ itself, based on gradient sign perturbation.
Note that the sign of the gradient is just $- \sign(\vw)$, and that $\vw^\top \sign(\vw) = ||\vw||_1$.
The adversarial version of
logistic regression is therefore to minimize
\[
    \mathbb{E}_{\vx, y \sim p_\text{data}} \zeta( y (\eps ||\vw||_1 - \vw^\top \vx - b)).
\]
This is somewhat similar to $L^1$ regularization. However, there are some important differences.
Most significantly, the $L^1$ penalty is subtracted off the model's activation during training, rather than
added to the training cost. This means that the penalty can eventually start to disappear if the model
learns to make confident enough predictions that $\zeta$ saturates. This is not guaranteed to happen---in
the underfitting regime, adversarial training will simply worsen underfitting.
We can thus view $L^1$ weight decay as being more ``worst case'' than adversarial training, because
it fails to deactivate in the case of good margin.

If we move beyond logistic regression to multiclass softmax regression, $L^1$ weight decay becomes
even more pessimistic, because it treats each of the softmax's outputs as independently perturbable,
when in fact it is usually not possible to find a single $\veta$ that aligns with all of the class's
weight vectors. Weight decay overestimates the damage achievable with perturbation even more in the
case of a deep network with multiple hidden units.
Because $L^1$ weight decay overestimates the
amount of damage an adversary can do, it is necessary to use a smaller $L^1$ weight decay coefficient than the $\eps$
associated with the precision of our features. When training maxout networks on MNIST, we obtained good results using
adversarial training with $\eps = .25$. When applying $L^1$ weight decay to the first layer, we found that even a coefficient of
.0025 was too large, and caused the model to get stuck with over 5\% error on
the training set. Smaller weight decay coefficients permitted succesful training
but conferred no regularization benefit.

\section{Adversarial training of deep networks}

The criticism of deep networks as vulnerable to adversarial examples is somewhat misguided, because
unlike shallow linear models, deep networks are at least able to {\em represent} functions that resist
adversarial perturbation. The universal approximator theorem~\citep{Hornik89} guarantees that a neural network
with at least one hidden layer can represent any function to an arbitary degree of accuracy so long as its hidden layer is permitted to
have enough units. Shallow linear models are not able to become constant near training points while
also assigning different outputs to different training points.

Of course, the universal approximator theorem does not say anything about whether a training algorithm
will be able to discover a function with all of the desired properties. Obviously, standard supervised
training does not specify that the chosen function be resistant to adversarial examples. This must be
encoded in the training procedure somehow.

~\citet{Szegedy-ICLR2014} showed that by training on a mixture of adversarial and clean examples, a
neural network could be regularized somewhat.
Training on adversarial examples is somewhat different from other data augmentation schemes; usually, one augments
the data with transformations such as translations that are expected to actually occur in the
test set. This form of data augmentation instead uses inputs that are unlikely to occur naturally
but that expose flaws in the ways that the model conceptualizes its decision function.
At the time, this procedure was never demonstrated to improve beyond dropout on a state of the art benchmark.
However, this was partially because it was difficult to experiment extensively with expensive adversarial examples
based on L-BFGS.

We found that training with an adversarial objective function based on the fast gradient sign method
was an effective regularizer: \[ \tilde{J}(\vtheta, \vx, y) = \alpha J(\vtheta, \vx, y) + (1-\alpha) J(\vtheta,
\vx + \eps \sign \left( \nabla_\vx J(\vtheta, \vx, y) \right). \]
In all of our experiments, we used $\alpha = 0.5$. Other values may work better; our initial guess
of this hyperparameter worked well enough that we did not feel the need to explore more.
This approach means that we continually update our supply of adversarial examples, to make them
resist the current version of the model.
Using this approach to train a maxout network that was also regularized with dropout, we were
able to reduce the error rate from 0.94\% without adversarial training to 0.84\% with adversarial
training.

We observed that we were not reaching zero error rate on adversarial examples on the
training set. We fixed this problem by making two changes. First, we made the model
larger, using 1600 units per layer rather than the 240 used by the original maxout network
for this problem. Without adversarial training, this causes the model to overfit slightly,
and get an error rate of 1.14\% on the test set. With adversarial training, we found that
the validation set error leveled off over time, and made very slow progress. The original
maxout result uses early stopping, and terminates learning after the validation set error
rate has not decreased for 100 epochs. We found that while the validation set error was
very flat, the {\em adversarial } validation set error was not. We therefore used early
stopping on the {\em adversarial validation set error}. Using this criterion to choose
the number of epochs to train for, we then retrained on all 60,000 examples. Five different
training runs using different seeds for the random number generators used to select minibatches
of training examples, initialize model weights, and generate dropout masks result in
four trials that each had an error rate of 0.77\% on the test set and one trial that had
an error rate of 0.83\%. The average of 0.782\% is the best result reported on the permutation
invariant version of MNIST, though statistically indistinguishable from the result obtained
by fine-tuning DBMs with dropout~\citep{dropout} at 0.79\%.

\begin{figure}
    \vspace{-.05in}
    \centering
    \begin{tabular}{cc}
        \includegraphics[width=.3\textwidth]{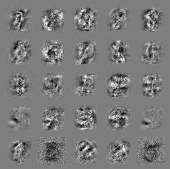} &
        \includegraphics[width=.3\textwidth]{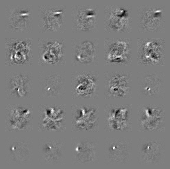}
    \end{tabular}
    \caption{Weight visualizations of maxout networks trained on MNIST. Each row shows
        the filters for a single maxout unit. Left) Naively trained model.
    Right) Model with adversarial training.}
    \label{fig:weights}
    \vspace{-.2in}
\end{figure}

The model also became somewhat resistant to adversarial examples. Recall that without
adversarial training, this same kind of model had an error rate of 89.4\% on adversarial
examples based on the fast gradient sign method. With adversarial training, the error rate
fell to 17.9\%. Adversarial examples are transferable between the two models but with the
adversarially trained model showing greater robustness. Adversarial examples generated via
the original model yield an error rate of 19.6\% on the adversarially trained model, while
adversarial examples generated via the new model yield an error rate of 40.9\% on the original
model. When the adversarially trained model does misclassify an adversarial example, its
predictions are unfortunately still highly confident. The average confidence on a misclassified
example was 81.4\%. We also found that the weights of the learned model changed significantly,
with the weights of the adversarially trained model being significantly more localized and
interpretable (see Fig.~\ref{fig:weights}).

The adversarial training procedure can be seen as minimizing the worst case error when the data is perturbed by an
adversary. That can be interpreted as learning to play an adversarial game, or as minimizing an
upper bound on the expected cost over noisy samples with noise from $U(-\eps, \eps)$ added to the
inputs. Adversarial training can also be seen as a form of active learning, where the model is
able to request labels on new points. In this case the human labeler is replaced with a heuristic
labeler that copies labels from nearby points.

We could also regularize the model to be insensitive to changes in its features that are smaller than the
$\eps$ precision simply by training on all points within the $\eps$ max norm box, or sampling many
points within this box. This corresponds to adding noise with max norm $\eps$ during training. However, noise with
zero mean and zero covariance is very inefficient at preventing adversarial examples. The expected dot product
between any reference vector and such a noise vector is zero. This means that in many cases the noise will have
essentially no effect rather than yielding a more difficult input. In fact, in many cases the noise will actualy result
in a lower objective function value. We can think of adversarial training as doing hard example mining among the set
of noisy inputs, in order to train more efficiently by considering only those noisy points that strongly resist
classification. As control experiments, we trained training a maxout network with noise based on randomly adding $\pm \eps$
to each pixel, or adding noise in $U(-\eps, \eps)$ to each pixel. These obtained an error rate of 86.2\% with confidence
97.3\% and an error rate of 90.4\% with a confidence of 97.8\% respectively on fast gradient sign adversarial examples.

Because the derivative of the sign function is zero or undefined everywhere, gradient descent on
the adversarial objective function based on the fast gradient sign method does
not allow the model to anticipate how the adversary will react to changes in the parameters.
If we instead adversarial examples based on small rotations or addition of the scaled gradient, then
the perturbation process is itself differentiable and the learning can take the reaction of the
adversary into account. However, we did not find nearly as powerful of a regularizing result from this
process, perhaps because these kinds of adversarial examples are not as difficult to solve.

One natural question is whether it is better to perturb the input or the hidden layers or both.
Here the results are inconsistent.
\citet{Szegedy-ICLR2014} reported that adversarial perturbations yield the best regularization when applied to
the hidden layers. That result was obtained on a sigmoidal network. In our experiments with the fast gradient
sign method, we find that networks with hidden units
whose activations are unbounded simply respond by making their hidden unit activations very large, so it is
usually better to just perturb the original input. On saturating models such as the Rust model we found that
perturbation of the input performed comparably to perturbation of the hidden layers.
Perturbations based on rotating the hidden layers solve the problem of unbounded activations growing to
make additive perturbations smaller by comparison. We were able to succesfully train maxout networks with
rotational perturbations of the hidden layers. However, this did not yield nearly as strong of a regularizing
effect as additive perturbation of the input layer.
Our view of adversarial training is that it is only clearly useful when the model has the capacity to
learn to resist adversarial examples. This is only clearly the case when a universal approximator theorem
applies. Because the last layer of a neural network, the linear-sigmoid or linear-softmax layer, is not
a universal approximator of functions of the final hidden layer, this suggests that one is likely to
encounter problems with underfitting when applying
adversarial perturbations to the final hidden layer. We indeed found this effect. Our best results with
training using perturbations of hidden layers never involved perturbations of the final hidden layer.
\section{Different kinds of model capacity}
One reason that the existence of adversarial examples can seem counter-intuitive is that most
of us have poor intuitions for high dimensional spaces. We live in three dimensions, so we are
not used to small effects in hundreds of dimensions adding up to create a large effect.
There is another way that our intuitions serve us poorly. Many people think of models with low
capacity as being unable to make many different confident predictions. This is not correct.
Some models with low capacity do exhibit this behavior. For example shallow RBF networks with
\vspace{-.05in}
\[
p(y=1 \mid \vx ) = \exp \left( (\vx - \mu)^\top \vbeta (\vx - \mu) \right) \]
are only able to confidently predict that the positive class is present in the vicinity of
$\mu$. Elsewhere, they default to predicting the class is absent, or have low-confidence
predictions.

RBF networks are naturally immune to adversarial examples, in the sense that they have low
confidence when they are fooled. A shallow RBF network with no hidden layers gets an error
rate of 55.4\% on MNIST using adversarial examples generated with the fast gradient sign
method and $\eps = .25$. However, its confidence on mistaken examples is only $1.2\%$.
Its average confidence on clean test examples is $60.6$\%.
We can't expect a model with such low capacity to get the right answer at all points of space,
but it does correctly respond by reducing its confidence considerably on points it does not
``understand.''

RBF units are unfortunately not invariant to any significant transformations
so they cannot generalize very well. We can view linear units and RBF units as different points on
a precision-recall tradeoff curve. Linear units achieve high recall by responding to every input in
a certain direction, but may have low precision due to responding too strongly in unfamiliar situations.
RBF units achieve high precision by responding only to a specific point in space, but in doing so
sacrifice recall. Motivated by this idea, we decided to explore a variety of models involving quadratic units,
including deep RBF networks. We found this to be a difficult task---very model with sufficient
quadratic inhibition to resist adversarial perturbation obtained high training set error when trained
with SGD.

\section{Why do adversarial examples generalize?}

An intriguing aspect of adversarial examples is that an example generated
for one model is often misclassified by other models, even when they
have different architecures or were trained on disjoint training sets.
Moreover, when these different models misclassify an adversarial example, they often
agree with each other on its class.
Explanations based on extreme non-linearity and overfitting cannot readily account
for this behavior---why should multiple extremely non-linear model with excess capacity
consistently label out-of-distribution points in the same way? 
This behavior is especially surprising from the view of the hypothesis that adversarial
examples finely tile space like the rational numbers among the reals, because in this
view adversarial examples are common but occur only at very precise locations.

\begin{figure}
    \centering
    \begin{tabular}{cc}
        \includegraphics[width=.4\textwidth]{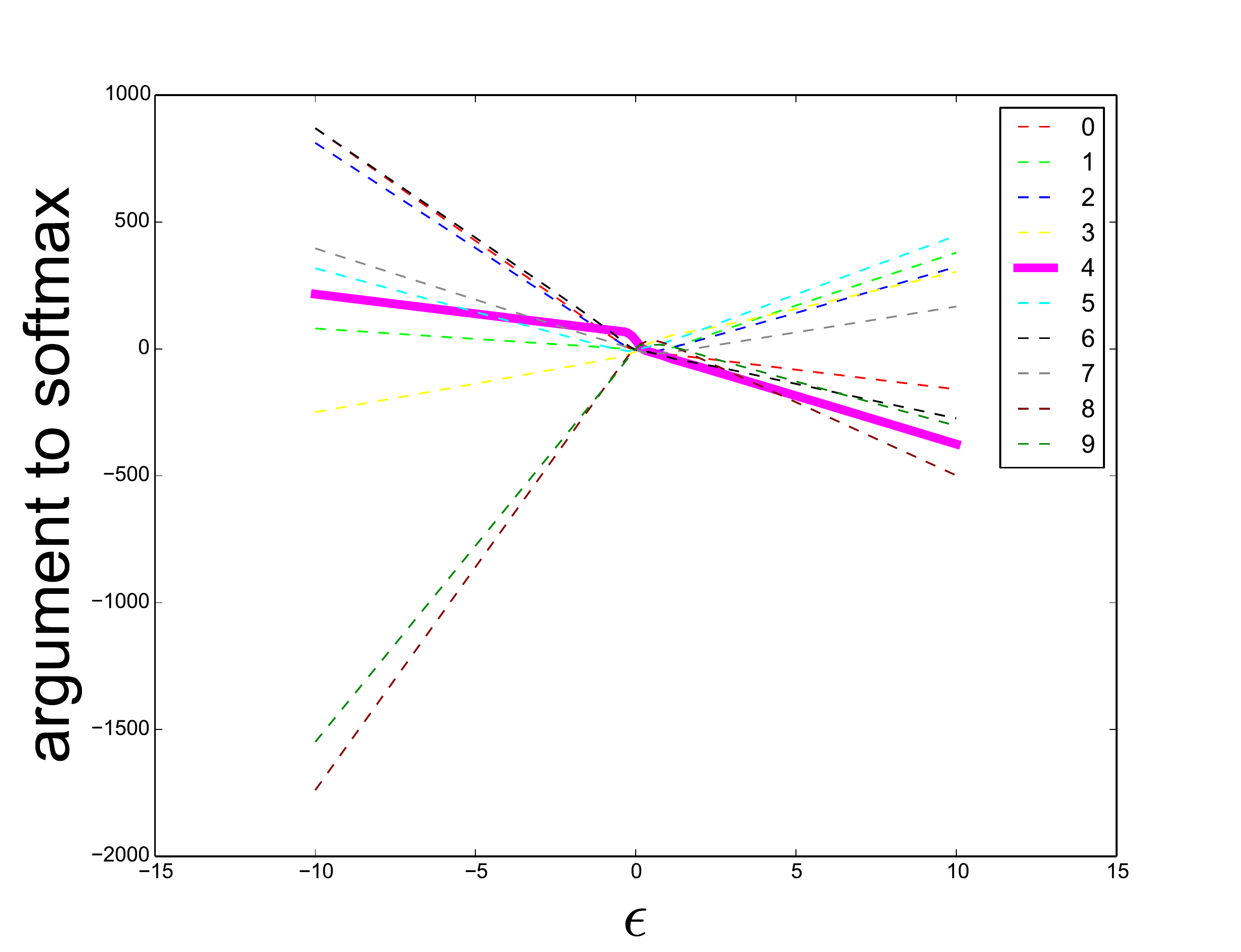} &
        \includegraphics[width=.3\textwidth]{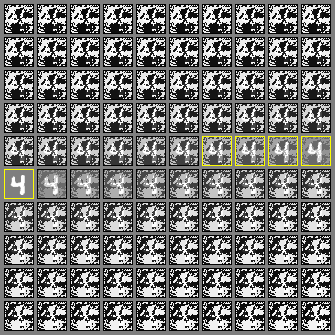}
    \end{tabular}
    \caption{
By tracing out different values of $\eps$, we can see that adversarial examples occur reliably
for almost any sufficiently large value of $\eps$ provided that we move in
the correct direction. Correct classifications occur only on a thin manifold where $\vx$ occurs in the data.
Most of $\mathbb{R}^n$ consists of adversarial examples and {\em rubbish class examples} (see the appendix).
This plot was made from a naively trained maxout network.
Left) A plot showing the argument to the softmax layer for each of the 10 MNIST classes as we vary $\eps$ on a single input example.
The correct class is 4. We see that the unnormalized log probabilities for each class are conspicuously piecewise linear with $\eps$ and that
the wrong classifications are stable across a wide region of $\eps$ values. Moreover, the predictions become very extreme as we
increase $\eps$ enough to move into the regime of rubbish inputs.
Right) The inputs used to generate the curve (upper left = negative $\eps$, lower right = positive $\eps$, yellow boxes indicate
correctly classified inputs).
    }
    \label{fig:eps_curve}
\end{figure}

Under the linear view, adversarial examples occur in broad subspaces. The direction $\veta$
need only have positive dot product with the gradient of the cost function,
and $\eps$ need only be large enough.
Fig.~\ref{fig:eps_curve} demonstrates this phenomenon. By tracing out different values
of $\eps$ we see that adversarial examples occur in contiguous regions of the 1-D subspace
defined by the fast gradient sign method, not in fine pockets. This explains why
adversarial examples are abundant and why an example misclassified by one classifier
has a fairly high prior probability of being misclassified by another classifier.

To explain why mutiple classifiers assign {\em the same class} to adversarial examples,
we hypothesize that neural networks
trained with current methodologies all resemble the linear classifier
learned on the same training set.
This reference classifier is able to learn approximately
the same classification weights when trained on different subsets of the training
set, simply because machine learning algorithms are able to generalize.
The stability of the underlying classification weights in turn results in the stability
of adversarial examples.

To test this hypothesis, we generated adversarial examples on a deep maxout network
and classified these examples using a shallow softmax network and a shallow RBF network.
On examples that were misclassified by the maxout network, the RBF network predicted
the maxout network's class assignment only 16.0\% of the time, while the softmax classifier
predict the maxout network's class correctly 54.6\% of the time. These numbers are largely
driven by the differing error rate of the different models though. If we exclude our
attention to cases where both models being compared make a mistake, then softmax regression
predict's maxout's class 84.6\% of the time, while the RBF network is able to predict maxout's
class only 54.3\% of the time. For comparison, the RBF network can predict softmax regression's
class 53.6\% of the time, so it does have a strong linear component to its own behavior.
Our hypothesis does not explain all of the maxout network's mistakes or all of the mistakes
that generalize across models, but clearly a significant proportion of them are consistent
with linear behavior being a major cause of cross-model generalization. 

\section{Alternative hypotheses}

We now consider and refute some alternative hypotheses for the existence of adversarial
examples. First, one hypothesis is that generative training could provide more constraint
on the training process, or cause the model to learn what to distinguish ``real'' from ``fake''
data and be confident only on ``real'' data.
The MP-DBM~\citep{mpdbm} provides a good model to test this hypothesis. Its inference procedure gets
good classification accuracy (an 0.88\% error rate) on MNIST. This inference procedure is differentiable.
Other generative models either have non-differentiable inference procedures, making it harder to compute
adversarial examples, or require an additional non-generative discriminator model to get
good classification accuracy on MNIST. In the case of the MP-DBM, we can be sure that the generative
model itself is responding to adversarial examples, rather than the non-generative classifier model
on top. We find that the model is vulnerable to adversarial
examples. With an $\epsilon$ of 0.25, we find an error rate of 97.5\% on adversarial examples generated
from the MNIST test set. It remains possible that some other form of generative training could
confer resistance, but clearly the mere fact of being generative is not alone sufficient.

%
Another hypothesis about why adversarial examples exist is that individual models have strange quirks but averaging over many models can
cause adversarial examples to wash out. To test this hypothesis, we trained an ensemble of twelve
maxout networks on MNIST. Each network was trained using a different seed for the random number generator
used to initialize the weights, generate dropout masks, and select minibatches of data for stochastic
gradient descent. The ensemble gets an error rate of 91.1\% on adversarial examples designed
to perturb the entire ensemble with $\epsilon = .25$. If we instead use adversarial examples designed to perturb only one
member of the ensemble, the error rate falls to 87.9\%. Ensembling provides only
limited resistance to adversarial perturbation.

\section{Summary and discussion}
As a summary, this  paper has made the following observations:
\begin{itemize}
\item Adversarial examples can be explained as a property of high-dimensional dot products. They are a result of models being too linear,
rather than too nonlinear.
\item The generalization of adversarial examples across different models can be explained as a result of adversarial perturbations being
highly aligned with the weight vectors of a model, and different models learning similar functions when trained to perform the same task.
\item The direction of perturbation, rather than the specific point in space, matters most. Space is not full of pockets of adversarial examples that finely tile the reals like the rational numbers.
\item Because it is the direction that matters most, adversarial perturbations generalize across different clean examples.
\item We have introduced a family of fast methods for generating adversarial examples.
\item We have demonstrated that adversarial training can result in regularization; even further regularization than
dropout.
\item We have run control experiments that failed to reproduce this effect with simpler but less efficient regularizers including $L^1$ weight
decay and adding noise.
\item Models that are easy to optimize are easy to perturb.
\item Linear models lack the capacity to resist adversarial perturbation; only structures with a hidden layer
(where the universal approximator theorem applies) should
be trained to resist adversarial perturbation.
\item RBF networks are resistant to adversarial examples.
\item Models trained to model the input distribution are not resistant to adversarial examples.
\item Ensembles are not resistant to adversarial examples.
  \end{itemize}
Some further observations concerning {\em rubbish class examples} are presented in the appendix:
\begin{itemize}
\item Rubbish class examples are ubiquitous and easily generated.
\item Shallow linear models are not resistant to rubbish class examples.
\item RBF networks are resistant to rubbish class examples.
\end{itemize}

Gradient-based optimization is the workhorse of modern AI. Using a network that has been designed to be sufficiently
linear--whether it is a ReLU or maxout network, an LSTM, or a sigmoid network that has been carefully configured not to saturate too much--
we are able to fit most problems we care about, at least on the training set. The existence of adversarial examples suggests that being
able to explain the training data or even being able to correctly label the test data does not imply that our models truly understand
the tasks we have asked them to perform. Instead, their linear responses are overly confident at points that do not
occur in the data distribution, and these confident predictions are often highly incorrect. This work has shown we can partially correct
for this problem by explicitly identifying problematic points and correcting the model at each of these points. However, one may
also conclude that the model families we use are intrinsically flawed. Ease of optimization has come at the cost of models that are
easily misled. This motivates the development of optimization procedures that are able to train models whose behavior is more locally
stable.

\subsubsection*{Acknowledgments}
We would like to thank Geoffrey Hinton and Ilya Sutskever for helpful discussions. We would also like to thank Jeff Dean, Greg Corrado, and Oriol Vinyals
for their feedback on drafts of this article.
We would like to thank the developers of Theano\citep{bergstra+al:2010-scipy,Bastien-Theano-2012}, Pylearn2\citep{pylearn2_arxiv_2013}, and DistBelief~\citep{distbelief}.

\small
\setlength{\bibsep}{4pt plus 0.3ex}
\bibliography{iclr2015}
\bibliographystyle{iclr2015}

\appendix
\section{Rubbish class examples}

A concept related to adversarial examples is the concept of examples drawn from a ``rubbish class.''
These examples are degenerate inputs that a human would classify as not belonging to any of the
categories in the training set. If we call these classes in the training set ``the positive classes,''
then we want to be careful to avoid false positives on rubbish inputs--i.e., we do not want to classify
a degenerate input as being something real.
In the case of separate
binary classifiers for each class, we want all classes output near zero probability of the class
being present, and in the case of a multinoulli distribution over only the positive classes, we would prefer that the
classifier output a high-entropy (nearly uniform) distribution over the classes.
The traditional approach to reducing
vulnerability to rubbish inputs is to introduce an extra, constant output to the model representing
the rubbish class~\citep{LeCun+98}.
\citet{fool} recently re-popularized the concept of the rubbish class in the context of computer vision
under the name {\em fooling images}. As with adversarial examples, there has been a misconception that
rubbish class false positives are hard to find, and that they are primarily a problem faced by deep networks.

Our explanation of adversarial examples as the result of linearity and high dimensional spaces
also applies to analyzing the behavior of the model on rubbish class examples. Linear models produce more extreme predictions at points that are
far from the training data than at points that are near the training data. In order to find high
confidence rubbish false positives for such a model, we need only generate a point that is far from
the data, with larger norms yielding more confidence. RBF networks, which are not able to
confidently predict the presence of any class
far from the training data, are not fooled by this phenomenon.

We generated 10,000 samples from $\mathcal{N}(0, \mI_{784})$ and fed them into various classifiers
on the MNIST dataset. In this context, we consider assigning a probability greater than 0.5 to any
class to be an error. A naively trained maxout network with a softmax layer on top had an error rate
of 98.35\% on Gaussian rubbish examples with an average confidence of 92.8\% on mistakes.
Changing the top layer to independent sigmoids dropped the error rate to 68\% with an average
confidence on mistakes of 87.9\%. 
On CIFAR-10, using 1,000 samples from $\mathcal{N}(0, \mI_{3072})$, a convolutional maxout net
obtains an error rate of 93.4\%, with an average confidence of 84.4\%.

These experiments suggest that the optimization algorithms employed by ~\citet{fool}
are overkill (or perhaps only needed on ImageNet), and that the rich geometric structure in
their fooling images are due to the priors encoded
in their search procedures, rather than those structures being uniquely able to cause false positives.

Though \citet{fool} focused their attention on deep networks, shallow linear models have the same problem.
A softmax regression model has an error rate of 59.8\%
on the rubbish examples, with an average confidence on mistakes of 70.8\%.
If we use instead an RBF network, which does not behave like a linear function,
we find an error rate of 0\%. Note that when the error rate is zero the average confidence on a mistake
is undefined.

\citet{fool} focused on the problem of generating fooling images for a specific class,
which is a harder problem than simply finding points that the network confidently classifies
as belonging to any one class despite being defective.
The above methods on MNIST and CIFAR-10 tend to have a very skewed distribution over classes.
On MNIST, 45.3\% of a naively trained maxout network's false positives were classified as 5s,
and none were classified as 8s. Likewise, on CIFAR-10, 49.7\% of the convolutional network's
false positives were classified as frogs, and none 
were classified as airplanes, automobiles, horses, ships, or trucks.

\begin{figure}
\centering
\includegraphics[width=0.5\textwidth]{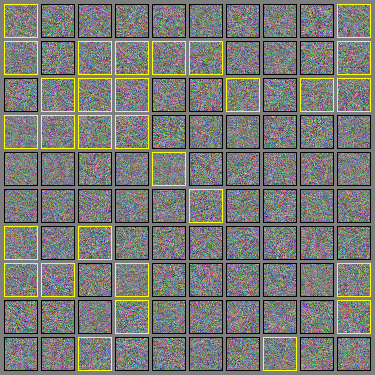}
\caption{
Randomly generated fooling images for a convolutional network trained on CIFAR-10.
These examples were generated by drawing a sample from an isotropic Gaussian, then
taking a gradient sign step in the direction that increases the probability of the
``airplane'' class. Yellow boxes indicate samples that successfully fool the model
into believing an airplane is present with at least 50\% confidence. ``Airplane'' is
the hardest class to construct fooling images for on CIFAR-10, so this figure
represents the worst case in terms of success rate.
}
\vspace{-.25in}
\label{fig:airplane}
\end{figure}

To solve the problem introduced by ~\citet{fool} of generating a fooling image for a particular
class, we propose adding $\eps \nabla_\vx p(y = i \mid \vx)$ to a Gaussian sample $\vx$ as
a fast method of generating a fooling image classified as class $i$. If we repeat this
sampling process until it succeeds, we a randomized algorithm
with variable runtime. On CIFAR-10, we found that one sampling step had a 100\% success rate for
frogs and trucks, and the hardest class was airplanes, with a success rate of 24.7\% per sampling
step. Averaged over all ten classes, the method has an average per-step success rate of 75.3\%.
We can thus generate any desired class with a handful of samples and no special priors,
rather than tens of thousands of generations of evolution. To confirm that the resulting examples
are indeed fooling images, and not images of real classes rendered by the gradient sign method,
see Fig.~\ref{fig:airplane}. The success rate of this method in terms of generating members of
class $i$ may degrade for datasets with more classes, since the risk of inadvertently increasing
the activation of a different class $j$ increases in that case.
We found that we were able to train a maxout network to have a zero percent error rate on Gaussian
rubbish examples (it was still vulnerable to rubbish examples generated by applying a fast gradient
sign step to a Gaussian sample)
with no negative impact on its ability to classify clean examples. Unfortunately, unlike
training on adversarial examples, this did not result in any significant reduction of the model's
test set error rate.

In conclusion, it appears that a randomly selected input to deep or shallow models built from
linear parts is overwhelmingly
likely to be processed incorrectly, and that these models only behave reasonably on a very
thin manifold encompassing the training data.

\end{document}